# ScheduleNanny: Using GPS to Learn the User's Significant Locations, Travel Times and Schedule

Parth Bhawalkar, Victor Bigio, Adam Davis, Karthik Narayanaswami, Femi Olumoko
College of Computing
Georgia Institute of Technology
Atlanta, GA 30332-0280 USA
{parthbh, vbigio, adamd, metlin, folumoko} @cc.gatech.edu

*Abstract - As computing technology becomes more pervasive, personal devices such as the PDA, cell-phone, and notebook should use context to determine how to act. Location is one form of context that can be used in many ways. We present a multiple-device system that collects and clusters GPS data into significant locations. These locations are then used to determine travel times and a probabilistic model of the user's schedule, which is used to intelligently alert the user. We evaluate our system and suggest how it should be integrated with a variety of applications.*

I. INTRODUCTION

Today's pervasive technology such as Personal Digital Assistants (PDAs), Cellular Phones and other similar tools have evolved from being viewed as mere organizers to be considered assistive technologies [1]. Users expect such tools to perform an increasingly large number of functions while remaining small and non-intrusive.

One of the primary requirements of any assistive technology is that it provide useful information to the user. This information needs to have immediate relevance to the tasks at hand. However, providing such contextual information is possible only with knowledge of the user's current context. For example, an alarm on a cell-phone or PDA is only helpful if that device has access to the current time of day.

In this paper, we describe a system that uses a particular context, location, in addition to time. We integrated off-the-shelf Global Positioning System (GPS) hardware with a PDA to collect location data. We implemented a system for processing this data to determine the user's significant locations, time of travel between such locations, and a probabilitic model of the user's weekly schedule. We use these results to intelligently alert the user of upcoming events on the PDA. Also, we provide an interface for the user to visualize the system's results and provide additional information.

*A. Previous Work*

In their paper [2], Daniel Ashbrook and Thad Starner describe how they use clustering to find signicant locations and create a predictive model of user movement. They used data for multiple people collected over an extended period of time with off-the-shelf GPS hardware. Although we use a similar method for clustering our data, they are interested in predicting where the user will go next based only on current location.

One system designed by Sparancino [3] used infared beacons to sense and predict user behavior. However, this system also concentrates on predicting user movement entirely as a function of user location.

There has been at least one system that has explored the idea of using the user's calendar as a sensor to gather user-location information, Mynatt and Tullio's Ambush [5]. This calendar system relied on user input to determine whether a user attended a meeting and concentrated mostly on prediction of user attendendance to future meetings and the sharing of such information.

*B. Intent of our Work*

As college students, we tend to have bad time management skills. Therefore, a device that learned our repeating weekly schedule and alerted us when we were running late for classes and other repeating events would be very useful. In addition, we are very busy, so we would like to give as little input as possible to this system.

To provide a solution to this problem, we decided to use a probabilistic model of a weekly schedule and alert the user when late for the next event. We also sought to provide the user with information on how long it would take to get to the next destination, based on the user's current location.

Since we sought to gather this information with minimal intervention from the user, we decided to investigate using only time and space information.

Here, we present a system that uses GPS data to create a probabilistic model of the user's location in time. We also extend this further by predicting approximate travel times between each significant location of the user's life. This information is correlated with our probabilistic model to intelligently alert the user.

II. APPLICATIONS

The ScheduleNanny system can relieve the user of guessing the time it takes to get to a destination and therefore aid in decidng when to depart in order to arrive at the next appointment on time. In this fashion the user can best utilize the time available.

The system has the potential to help a user take advantage of opportunities, and thereby increase the user's efficiency. When combined with a to-do list the system can use known places to remind the user of pending tasks at or near the current location.

Since the system records continuous track of the user's movements through space and time, the data can easily be used to provide a log of when the user went where and the time it took him/her. In fact, the movements of a given user can be mapped and a simulation of the user's movements can be generated.

On a collaborative level, the system can be used to help answer two questions: 1) where is the user at this time, and 2) where is the user likely to be in the future. Both of these can be used to aid a colleague in locating and or meeting with the user, as in previous work [5].

### III. METHODS

The various sections of our system can be broadly classified as the data gathering module (composed of the GPS and the PDA), the machine learning module (performed on the PC) and the alerting module (the PDA).

*A. System Design*

The PDA and the GPS receiver implement the primary input unit, and are used to gather information about the user's movements through space and time. The software on the PDA samples the GPS receiver, and stores valid tracking information (latitude, longitude, time) once per second. Tracking information is uploaded from the PDA to the PC on a daily basis.

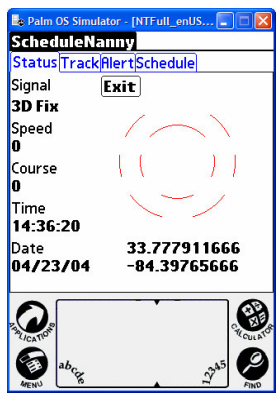

**Figure 1. Screen shot of the GPS status screen of our Palm application. This shows the user the operation status of the physical GPS tracker.**

The machine learning module of the system is contained in the user's PC. The tracking data from the PDA is integrated with previous tracking data, and the machine learning routines use time and location information to 1) determine the user's significant places, 2) calculate travel times between these places, and 3) to generate a probabilistic schedule (mschedule) for the user. These three tables (places, travel times, mschedule) are then transferred to the PDA on a daily basis.

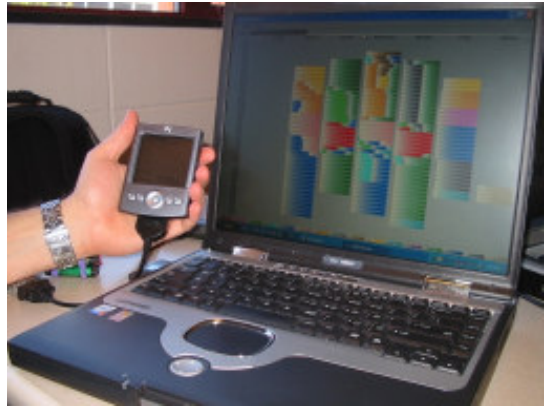

**Figure 2. Synchronizing the PDA with the PC.**

The PDA software uses the tables generated by the machine learning routines to alert the user if he/she will be late for an appointment. Every 15 seconds the system examines the appointments that start within the next two hours, and then calculates the travel time from the current location to the appointment location. The current time, travel time and appointment time are used to determine whether the user might be late to this appointment, and alerts the user if this is the case.

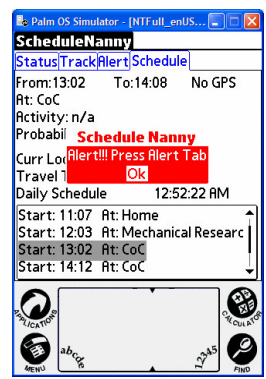

**Figure 3. Screen shot of an alert produced on the Palm application. The user is prompted to select the Alert Tab, which shows further details about the alert.**

## B. Architecture

Figure 4 broadly illustrates the architecture of our system. The user's movements through space and time (GPS data) are fed into our machine learning (ML) filter. The ML filter uses clustering techniques to find the user's significant places (where he/she spends time). Travel times are calculated by using the users movement times from place to place. The system generates a probabilistic schedule based on the times that the user is at his/her significant locations and the frequency with which this occurs. An alert is generated when the user's travel time from the current location will exceed the start time of a given appointment in the schedule.

Although not implemented, the user's calendar is intended to inform the probabilistic schedule, and should help the system assign names to significant places, assign activities to appointments, and to achieve a higher certainty when generating the probabilistic schedule.

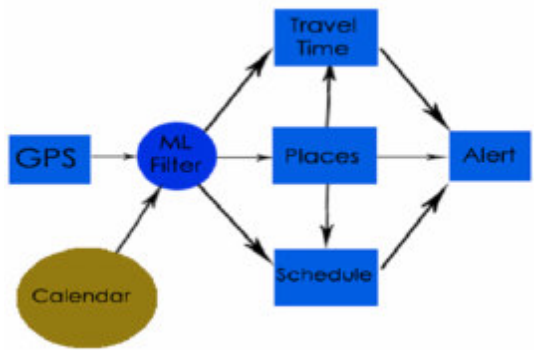

**Figure 4. High-level model of the ScheduleNanny architecture.**

## C. Data Processing - Clustering

When clustering our GPS points into significant locations, we use a method similar to that of Ashbrook and Starner [2].

First, the acquired GPS data is preprocessed by removing all points with speed less than one mile per hour. This left us with only those points when the user is in motion. Next, we find all the points that precede a gap of at least T minutes. We can assume such a point is a significant location with some confidence, since the user did not receive GPS signal, or did not move for an extended period of time. Any time could arbitrarily be picked for T; however, we chose T to be 10 minutes, based on the results observed by Ashbrook and Starner. This procedure leaves us with the set of points that are of interest to our system. We call these points *places*.

Since GPS measurements are not completely accurate, simply using the places would not give meaningful results. For this reason, we cluster these points using a modified k-means algorithm. We call each resulting cluster a *location*, which we define as a circle in space with a center and radius. To further simplify our approach, we assume all locations have the same radius. Since the radius picked for our locations should be adaptive to the user, we must decide adaptively what radius to choose.

To select the best radius, we do a search over a set of possible radii. For each size of radius, we cluster all the data points and count the total number of clusters. To find the clusters, we use a modified k-means approach, starting with a random cluster and iteratively refining it. After all the points have been clustered, we remove any clusters owning no unique points. We then store the number of clusters associated with the given radius. After all the radii have been considered, we can plot the number of clusters by radius size in a graph. Figure 6 shows a sample of this graph for GPS data that we acquired.

In previous work by Ashbrook and Starner, a particular knee in the graph was found. They used this method recursively to cluster the data into locations and sub-locations. However, we found this unnecessary for our approach, since we are interested in locations the user will associate with significant events in her life. Instead, we first smoothened the graph as shown in Figure 7, which removed the noise due to the randomness of our algorithm. Next, we moved from right to left and chose the radius where the derivative of the graph essentially stopped decreasing.

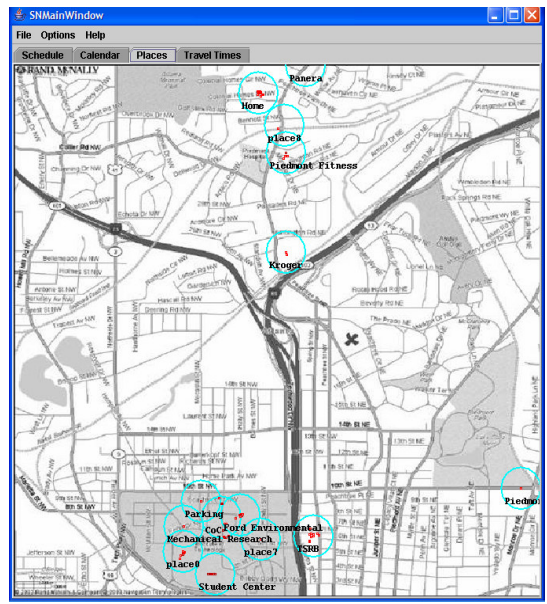

**Figure 5. Screenshot of our clustered locations projected onto a map of Atlanta, GA.**

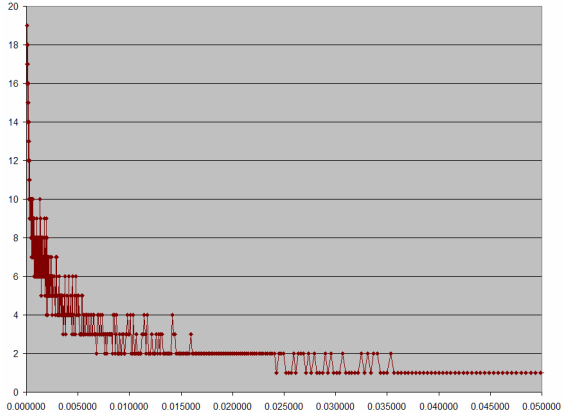

**Figure 6. Graph of number of clusters by the size of the radius.**

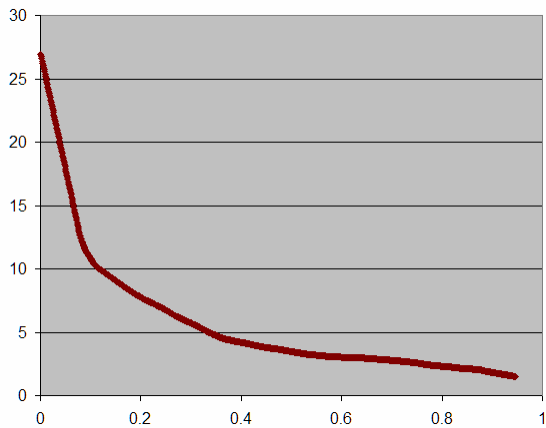

**Figure 7. Smoothened graph of number of clusters by the size of the radius.**

*D. Data Processing – Creation of the Schedule*

We define the *Schedule* as a mapping from a time and location to a probability for any given minute of a day of the week. This probability represents the likelihood of the user being at the given location at the given time:

$$\text{Schedule}(T, L) \rightarrow [0,1]$$

In the above equation, *T* can be any minute of any weekday and *L* is a location found by our clustering algorithm.

To simplify our approach, we made the schedule equal to a statistical model of the location of the user for each minute of every day. The learned significant locations are used to locate the user at any point in time. The method used in creating the schedule is identical to that used in place clustering, except the user's location at each time segment is extrapolated and stored.

However, finding the start times and end times proved to be difficult since we do not usually reacquire the user's GPS signal until long after the event has ended. Therefore, we calculate the travel time between events to estimate the end time. Once we have the travel time, the end-time of the event is simply the beginning time of the next event minus the travel time:

Event time = beginning of next event – travel time

So, the travel times fulfill two purposes: determining the user's schedule, and calculating when to alert the user for the next appointment.

*E. Data Processing – Calculation of Travel Times*

To calculate the travel times between each known location, we consider every instance when the user leaves one event and travels to the next event. Since we define an event as a stopping point of more than 10 minutes, this ensures that our travel times will generalize. The average times of travel between pairs of locations are then collected. Since the GPS signal is noisy and broken during the user's travel, calculating each travel time is not trivial.

To overcome the gaps in our data, we exploit the information we do have. Since we know where the user started and ended as GPS points we have the distance. Also, we have some estimate of the user's average speed. So, we calculate travel time as the following:

$$T = \text{Distance}(AB) / \text{Speed}(BC) + \text{Time}(BC).$$

Where A is the starting location, C is the ending location, and B is the first data point aquired after A with speed above 1mph.

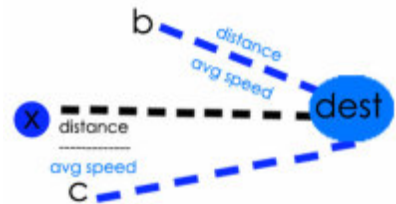

**Figure 8. Scenario of calculating a travel time for the user on-the-fly.**

When the user is in an unknown location, the travel time to the user's next possible appointment is also unknown and must be calculated on-the-fly by the PDA.

We used a simple method of calculation of these travel times that accurately estimates actual travel times.

If X is the user's current location, the travel time *T* is calculated:

*T* = Distance(X to destination) / Speed(most efficient known path)

Where the most efficient known path is defined as the fastest path through a known location from X to the destination. The following scenario is shown in Figure 8: the user is at point X and locations b, c, and dest are the only known locations, where dest is the destination and (X, c, dest) is the most efficient known path.

*F. Visualizing the Schedule*

Visualizing a statistical model of the user's past location is a difficult problem that has been tackled before [5]. We choose to split each hour of each day into multiple segments as specified by the user. Each segment contains a block colored to represent a location. The location colors are shown in a legend at the bottom of the window. Each block's size reflects the percentage the user was at that location at that time; so, only a few blocks appear for a given time segment. To find the values for each time segment, we simply took the average over that time segment. Figure 9 shows our representation of a schedule with time segments of 30 minutes.

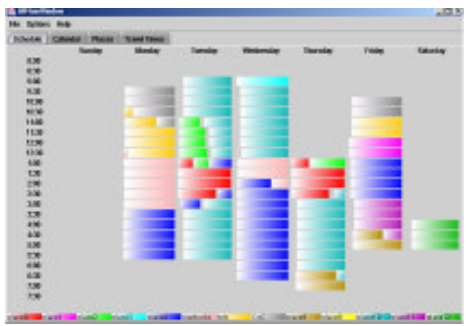

**Figure 9. A sample user schedule as displayed by our PC application.**

IV. TECHNOLOGY

Time and budgetary constraints required us to look for a solution that could be implemented with readily available components and could be developed rapidly enough to allow appropriate time for testing.

Our goal was to build a system that is easily built from off the shelf components, and can be tested across multiple users without having to buy expensive equipment. We also wanted to design the software for the system in a language that was easy to port and duplicate, and would work across multiple platforms without requiring too much effort of customization.

*A. Production System*

GPS information is gathered using a standard NMEA-0183 compatible GPS receiver connected to a Palm Tungsten T. The same Palm PDA is used to alert the user if he/she sill be late for an appointment. The PC system used to process the GPS data and run the machine learning algorithms is a Compaq Evo N800v with 256MB RAM running at 2GHz. A variety of specialized cables are required to connect GPS to Palm, GPS to PC, and Palm to PC.

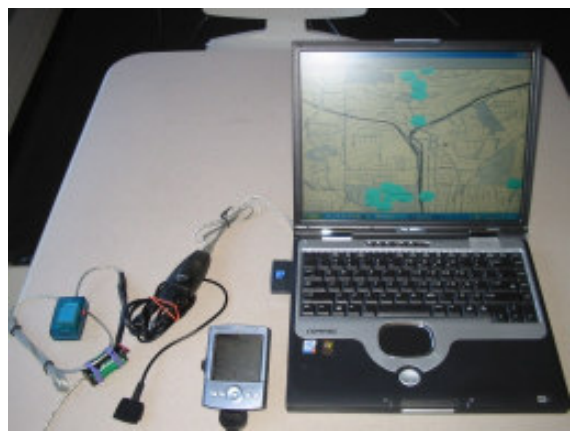

**Figure 10. The ScheduleNanny hardware.**

*B. Development systems*

Cetus GPS software was initially deployed on the PDA to gather GPS tracking information. The PDA-based applications were developed using SuperWaba, a Java equivalent language that permits rapid development for portable devices. The Palm Simulator, Tauschke MobileCreator and Apache Ant were used in conjuction with SuperWaba. The machine learning application was developed in a Windows XP based desktop using Sun's Java.

V. EVALUATION

We conducted preliminary evaluations of the system prototype to test the usability of the system software and the usefulness of the system to an eventual user.

*A. Usability*

Cognitive Walkthrough and Heuristic evaluation techniques were employed to perform the initial evaluations of the PDA and PC software interfaces. The evaluation consisted of four HCI experts who were aware of common Graphical User Interface (GUI) usability issues and had experience conducting usability evaluations. None of the chosen experts had any prior exposure to the system, therefore a brief system description and demonstration was given prior to each evaluation session.

V.A.1    Cognitive Walkthrough

We were interested in evaluating the "learnability" of the software interface and uncovering any usability issues that threaten effective user interaction with the system (especially novice users). Two expert walkthroughs of both interfaces (PDA and PC) were conducted. Each expert was initially presented with a script containing three tasks to be preformed on the system, together with the required actions to complete the tasks. The tasks chosen were typical tasks we expect eventual users of the system to perform. The tasks include the following:

1. Adding an appointment to the user's calendar
2. Modifying a current calendar entry to reflect a location change.
3. Determining if an alert had been missed within the past few hours.

Each expert was instructed to perform the task (by stepping through the action sequence provided) while noting the potential usability issues they came across.

V.A.2    Heuristic Evaluation

We were also interested in evaluating how the system conformed to some standard usability criteria. Consequently, two expert evaluations were performed whereby the system was carefully critiqued against a given set of heuristics. Prior to the evaluation, the experts were presented with a script containing a list of seven of the ten common heuristics [6] used to asses the usability of a Graphical User Interface (GUI). The experts were encouraged to note any usability problem encountered without worrying if it fit under a given heuristic or not.

The chosen heuristics are listed below together with our justification for choosing them.

  i. Visibility of System Status: To ensure that users were always informed about what was going on in the system, and that the feedback was appropriate and reasonable.
 ii. Match between System and Real World: To ensure that the terminology and concepts used would be familiar to the user.
iii. Recognition rather than Recall: To ensure that all the available actions/options were visible and the appropriate information was available on the screen when needed.
 iv. Consistency: To ensure that the actions/operations and terminology remained consistent throughout the interface.
  v. User Control and Freedom: To ensure that users always felt in control of the system.
 vi. Simple Dialog: To ensure that irrelevant or rarely needed information was not present and did not detract from other important information.
vii. Error Prevention, Diagnosis and Recovery: To ensure that the interaction was not error prone and that adequate provisions were made for detection, diagnosis and recovery form the error.

*B. Usefulness*

It was equally important to evaluate the usefulness of the system, in particular, its accuracy and effectiveness. This involved a user "living" with the system for a significant period of time. Due to time and resource constraints (availability of hardware), we could only perform the evaluation with one user (member of the system's development team) who lived with the system for about two weeks.

Following the two week evaluation period, a questionnaire was then deployed to obtain the user's assessment of the accuracy of the system's appointment alerts and travel time estimates, its portability, ease of use, as well as the user's overall trust and satisfaction with the system. The user was also encouraged to give accounts (if any) of any effect the system had on his daily behavior.

VI.    RESULTS

The results of our evaluation methodologies were qualitative and allowed us to ascertain the benefits of using the system. We needed to understand problems in our GUI since the system was spread across two platforms. We summarize some of the results of our evaluation both with the experts and the user.

| Heuristic | Severity Rating |
|---|---|
| Visibility | 3 |
| Match between system and real world | 3 |
| Recognition rather than recall | 4 |
| Consistency | 2 |
| User Control and Freedom | 2 |
| Minimalist design | 1 |
| Error Prevention. | 3.5 |

**Table 1. Heuristic evaluation severity ratings.**

### A. Heuristic Evaluation & Cognitive Walkthrough

The experts assigned severity ratings to our chosen set of usability principles. The results are summarized in the table. The severity ratings are based on the Likert scale, where 0 means no problems with the interface and 4 means serious design flaws.

Both the heuristic evaluation and cognitive walkthrough uncovered similar usability issues. The most significant issue the experts found was the inappropriate labeling and lack of visibility in the elements of the system software, which made the interface less intuitive. Another significant usability issue that the experts pointed out was the absence of any undo functionality which violated the error prevention and recovery usability heuristic. The experts suggested that our system was over-specific in representing the travel times, as the precision was up to 5 decimal places, which was not particularly useful and hard to read. The visualization technique used to represent the probabilistic schedule was also found to be complex and had some readability issues.

One final usability problem uncovered was the lack of freedom and control provided to the user. More specifically, the experts noted that too much effort was required by the user to add a calendar item or modify one.

The usability problems were mostly cosmetic in nature and did not significantly affect the functioning of the system.

### B. Living with the system

As described in the previous section, we evaluated the system for usefulness and had a member of the team live with the system. Based on feedback from the user through the questionnaire, we found out that the system provided a very good estimate of the travel time between locations. Occasionally the system alerted about future appointments it thought the user would be late for. However, lack of time prevented us from truly knowing if the system was consistent in its alerts.

### VII. CONCLUSIONS

We defined our goal as getting a user to his destination on time. After gathering over two months of GPS data, we feel that we are in a position to answer some of our original questions. Using the user's movements through space and time (GPS) we've learned what locations are significant to a user, and the travel times between them. We can develop a probabilistic model of the user's location for each minute of the day and we can create a probabilistic schedule.

In actual use we discovered that though the system can learn pertinent facts regarding where the user spends time, a probabilistic model of the users schedule is only sufficient to inform him where he is likely to be. It is not enough to ascertain where he should be.

We purposefully set out to see how far we could get with just information gathered with no explicit user input and see if we could predict the user's schedule. We think that the system will be more effective if we use the places and travel times to inform a user's calendar and the calendar to constrain the user's schedule.

### VIII. FUTURE WORK

We have created a framework which can be applied to other applications within the same domain. One possible extension would be to predict where a user should be.

We would also like to integrate our system with a host of information gathering agents such as email reading programs, gaze trackers, traffic and weather information sources and other context-aware agents to increase the accuracy and timing of the schedule and alerts. Also, our system could be integrated with calendar assistants, for example, see [5].

Apart from the applications that the Schedule Nanny can be extended to, we need to evaluate the system over a larger time period and with a larger set of people. In the future, with further iterations through the interface development cycle, we plan to perform more usability evaluations of the PDA and PC software interfaces. In future evaluations, we anticipate each user maintaining an alert log throughout the evaluation period. This data will later be used to judge the accuracy and effectiveness of the appointment alerts.

### ACKNOWLEDGMENTS

The authors thank Professors Charles Isbell and Jeff Pierce of the GVU Center, Georgia Institute of Technology. We also thank Professor Thad Starner and Dan Ashbrook at Georgia Institute of Technology for providing us with the necessary hardware for this project.